\documentclass[sigconf]{acmart}
\AtBeginDocument{%
  }

\setcopyright{acmlicensed}
\copyrightyear{2025}
\acmYear{2025}
\acmDOI{XXXXXXX.XXXXXXX}
\acmConference[Prompt Optimization KDD 2025]{August 4--5, 2025}{Toronto, Canada}

\usepackage{xcolor}
\usepackage{amsmath}
\usepackage{algorithm}
\usepackage{algorithmic}
\usepackage[parfill]{parskip}
\usepackage{multirow}
\usepackage{booktabs}

\usepackage{listings}
\usepackage{xcolor}

\definecolor{codegreen}{rgb}{0,0.6,0}
\definecolor{codegray}{rgb}{0.5,0.5,0.5}
\definecolor{codepurple}{rgb}{0.58,0,0.82}
\definecolor{backcolour}{rgb}{0.95,0.95,0.92}

\lstdefinestyle{mystyle}{
    backgroundcolor=\color{backcolour},   
    commentstyle=\color{codegreen},
    keywordstyle=\color{magenta},
    numberstyle=\tiny\color{codegray},
    stringstyle=\color{codepurple},
    basicstyle=\ttfamily\footnotesize,
    breakatwhitespace=false,         
    breaklines=true,                 
    captionpos=b,                    
    keepspaces=true,                 
    numbers=left,                    
    numbersep=5pt,                  
    showspaces=false,                
    showstringspaces=false,
    showtabs=false,                  
    tabsize=2
}

\usepackage{soul}

\begin{document}

\title{Tournament of Prompts: Evolving LLM Instructions Through Structured Debates and Elo Ratings}

\author{Anirudh Nair\textsuperscript{*}}
\affiliation{%
  \institution{Amazon}
  \city{Boston}
  \state{MA}
  \country{USA}}
\email{rianina@amazon.com}

\author{Adi Banerjee\textsuperscript{*}}
\affiliation{%
  \institution{Amazon}
  \city{New York}
  \state{NY}
  \country{USA}}
\email{adibaner@amazon.com}

\author{Laurent Mombaerts}
\affiliation{%
  \institution{Amazon}
  \city{Luxembourg}
  \country{Luxembourg}}
\email{lmomb@amazon.com}

\author{Matthew Hagen}
\affiliation{%
  \institution{Amazon}
  \city{Atlanta}
  \state{GA}
  \country{USA}}
\email{mathage@amazon.com}

\author{Tarik Borogovac}
\affiliation{%
  \institution{Amazon}
  \city{Boston}
  \state{MA}
  \country{USA}}
\email{tarikbo@amazon.com}

\renewcommand{\shortauthors}{Nair and Banerjee et al.}

\begin{abstract}

Prompt engineering represents a critical bottleneck to harness the full potential of Large Language Models (LLMs) for solving complex tasks, as it requires specialized expertise, significant trial-and-error, and manual intervention. This challenge is particularly pronounced for tasks involving subjective quality assessment, where defining explicit optimization objectives becomes fundamentally problematic. Existing automated prompt optimization methods falter in these scenarios, as they typically require well-defined task-specific numerical fitness functions or rely on generic templates that cannot capture the nuanced requirements of complex use cases. We introduce DEEVO (DEbate-driven EVOlutionary prompt optimization), a novel framework that guides prompt evolution through a debate-driven evaluation with an Elo-based selection. Contrary to prior work, DEEVO's approach enables exploration of the discrete prompt space while preserving semantic coherence through \textit{intelligent crossover} and \textit{strategic mutation} operations that incorporate \textit{debate-based feedback}, combining elements from both successful and unsuccessful prompts based on identified strengths rather than arbitrary splicing. Using Elo ratings as a fitness proxy, DEEVO simultaneously drives improvement and preserves valuable diversity in the prompt population. Experimental results demonstrate that DEEVO significantly outperforms both manual prompt engineering and alternative state-of-the-art optimization approaches on open-ended tasks and close-ended tasks despite using no ground truth feedback. By connecting LLMs' reasoning capabilities with adaptive optimization, DEEVO represents a significant advancement in prompt optimization research by eliminating the need of predetermined metrics to continuously improve AI systems.

\end{abstract}

\thanks{\textsuperscript{*}These authors contributed equally to this work}







\maketitle

\section{Introduction}
Large Language Models (LLMs) have demonstrated remarkable capabilities across diverse domains, such as literary and professional writing, code generation, and problems requiring logical reasoning. However, their performance towards a specific task remains heavily dependent on the quality of instructions — or prompts — provided to them \cite{fernando2023promptbreeder, guo2025evopromptconnectingllmsevolutionary}. The term \textit{prompt engineering} has become widely used, signaling that prompting has become an important skill for harnessing these models' full potential. This skill requires specialized expertise attained through learning techniques and significant trial-and-error. Furthermore, when prompts prove insufficient for a task, developers must either implement dedicated post-processing logic or employ fine-tuning strategies to address performance gaps. In that sense, the development of systems executing complex tasks while solely relying on prompt engineering is resource intensive, thus motivating the need for an automated method to optimize prompts.

The automated optimization of prompts is especially challenging for tasks where performance or quality is judged subjectively, where ambiguity challenges require resolution, or where managing conflicting contexts is paramount \cite{suzgun2022challenging}. In these scenarios, agents are expected to learn to adapt synthesizing different types of information, make judgment calls from different perspectives, and self-ascertain branching and stopping conditions during its iteration process; all in the absence of any quality criteria or scoring functions to quantify success along said criteria.  


Current approaches to automated discrete prompt optimization fall into two primary categories: gradient-based methods and evolutionary strategies. Gradient-based methods operate on textual gradients defined by means of LLM generated critical feedback (examples of these are Protegi \cite{pryzant2023automaticpromptoptimizationgradient} and TextGrad \cite{yuksekgonul2024textgradautomaticdifferentiationtext}). These methods offer computational efficiency but typically require labeled ground truth data from which to calculate loss; risk task-specific overfitting especially when there is a lack of diversity in the examples; and do not have a mechanism to perform exploration and thus suffer from adaptability issues. Conversely, evolutionary methods (such as EvoPrompt \cite{guo2025evopromptconnectingllmsevolutionary} and PromptBreeder {\cite{fernando2023promptbreeder}}) provide broader exploration capabilities but suffer from computational inefficiency due to random search and, crucially, depend on well-defined objective fitness functions — which are often unavailable for subjective tasks.

We introduce DEEVO (DEbate-driven EVOlutionary prompt optimization), a novel framework that addresses the challenges of 1) exploration vs exploitation, and 2) lack of labeled ground truth / fitness functions, by guiding prompt evolution through structured debates and Elo-based selection.

\textbf{Exploration vs Exploitation}
\newline Unlike previous approaches, DEEVO enables systematic exploration of the prompt space through two innovative evolutionary mechanisms. At its core, DEEVO employs multi-agent debate \cite{du2023improving} to guide intelligent crossover, where strengths and weaknesses of parent prompts are identified before strategically combining their effective elements. This process is complemented by targeted prompt mutations that specifically modify instructions to improve task performance. DEEVO's evolutionary approach selectively incorporates elements from both successful and unsuccessful prompts based on their identified strengths, preserving prompt effectiveness and logical structure while systematically exploring the solution space.


\textbf{Lack of Labeled Ground-Truth Data}
\newline Unlike prior prompt optimization strategies, DEEVO does not rely on labeled data / fitness functions against ground truth in order to evaluate prompt effectiveness. Instead, it leverages LLM-powered multi-agent debates (MAD) \cite{du2023improving} to evaluate prompt quality without requiring predetermined metrics. By having LLMs critique prompt outputs in a pairwise fashion and determine a winner through structured debates, DEEVO ensures a self-contained evaluation system that can assess quality across diverse tasks — including those with subjective criteria. The evaluation mechanism evolves and improves along with the prompts, incorporating novel ideas as candidate criteria for future evaluation. The resulting debate verdicts serve as a fitness proxy that simultaneously drives improvement and preserves valuable diversity in the prompt population without needing a manually crafted or separately learned objective function for fitness selection.

\textbf{A Modified Elo-Based Selection}
\newline The Elo rating system \cite{elo1978rating} is a robust method to rank entities (in this case, prompts) via pairwise comparisons, where each prompt maintains a numerical rating that dynamically updates based on competition outcomes. This approach has gained significant traction in LLM evaluation frameworks, with numerous benchmark systems adopting Elo-based mechanisms to rank model and prompt performance \cite{ameli2024statisticalframeworkrankingllmbased, chiang2024chatbotarenaopenplatform, boubdir2023elouncoveredrobustnessbest}. Despite its robustness, Elo is particularly limited in its ability to handle newcomer prompt competitors (due to requiring many matchups to reach an accurate skill assessment); and veteran prompt competitors (due to their ratings becoming "sticky" over time, as historical matchups dilute recent performances). For this reason, DEEVO utilizes a modified Elo-selection mechanism that introduces selection quotas for newcomers and veterans; to force the prompt population to always consist of a balanced proportion of newcomer and veteran prompts. This enforces a weaker barrier of entry to new prompt candidates, better captures current skill levels for veterans, and provides more accurate performance estimates. 

We demonstrate that DEEVO significantly outperforms both manual prompt engineering and alternative optimization approaches across both open and close ended tasks. 
By connecting LLMs' reasoning capabilities with adaptive optimization, DEEVO eliminates the need for developing predetermined metrics to continuously optimize prompts, opening new possibilities for self-improving AI systems across domains where subjective quality assessment is essential. 


\begin{figure*}
  \includegraphics[width=\textwidth]{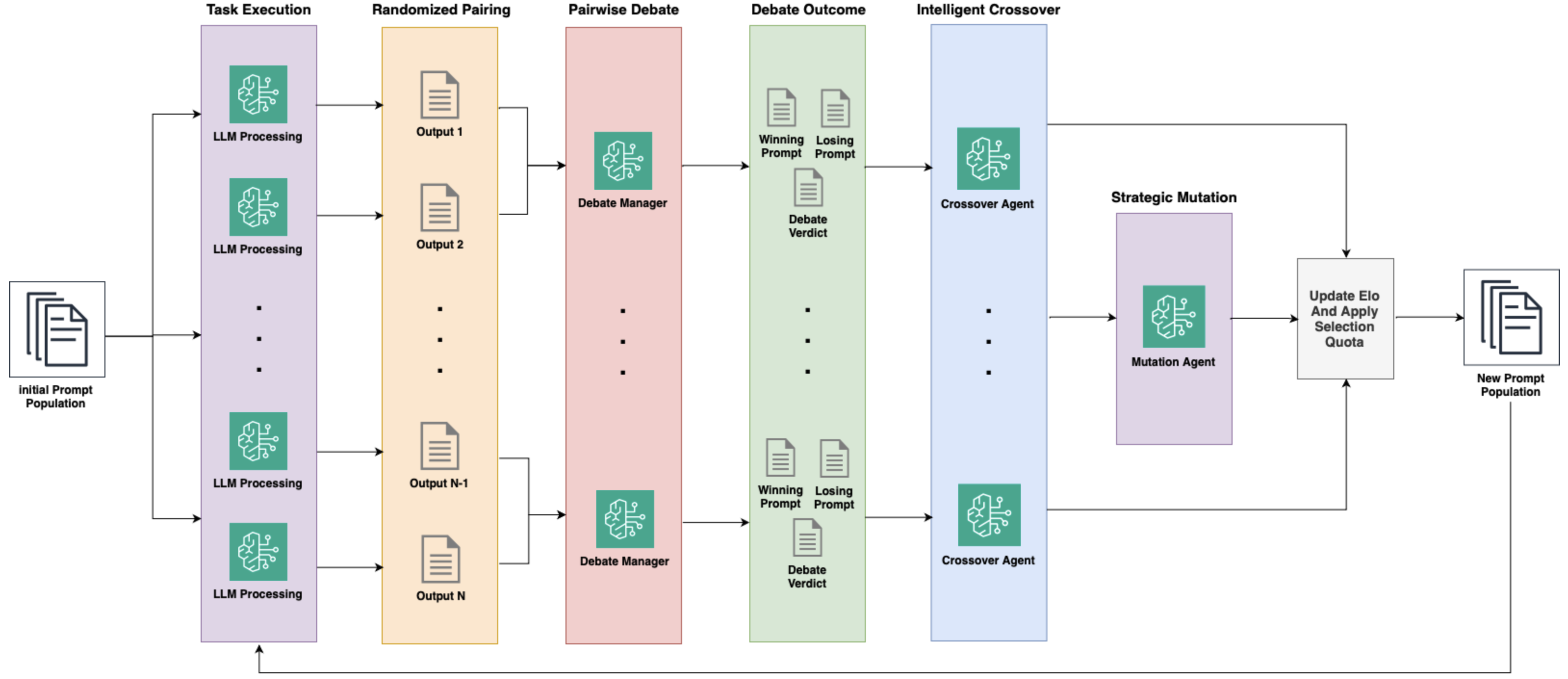}
  \caption{DEEVO. (1) First an initial set of prompts are either provided or generated. (2) Next, each prompt is executed. (3) Each output is then is then paired with another and then passed into a Multi-Agent Debate evaluation to determine a winner. (4) A Crossover agent then leverages the debate trace to intelligently create a new prompt that combines the strong elements of both prompts in each pair. (5) Some child prompts are then randomly selected to go through a task-driven Mutation agent. (6) Finally, the Elo ratings are updated based on the winner and loser prompts while children are given a base rating of 1000, and the next generation repeats.}
  \label{fig:deevo}
\end{figure*}

\section{Related Works}
\subsection{Prompt Optimization}
Prompt optimization has emerged as a critical area of research in large language model (LLM) development, with researchers exploring various techniques to enhance model performance through systematic refinement of input prompts. Historically, researchers have relied on manual supervised approaches, using black-box techniques that score prompts based on observable output metrics such as accuracy, F1 score, BLEU, or ROUGE. However more recently, automatic prompt optimization has come into light as an alternative, scalable solution \cite{shin2020autoprompt}. Soft prompt optimization operates in a continuous space and usually involves some gradient-based operator. These methods operate in a continuous space range by leveraging embeddings to automatically optimize the prompts \cite{zhang2021differentiable, sun2022black, hu2024localized, NEURIPS2024_061d5d1b}, training auxiliary models to output optimized prompts \cite{jain2024prompt, cheng2023black, deng2022rlprompt, sun2024prompt, zhang2023tempera}, or using non-gradient approaches to adjust prompt representations \cite{chen2023instructzero}. 

In spite of effective performance, continuous methods often lack interpretability \cite{lester2021power}, require model training \cite{zhu2024bdpl}, or need access to at least partial knowledge of the internals of an LLM \cite{prasad2022grips}, something out of scope for black-box LLM APIs. Contrary to optimizing in a continuous space, discrete prompt optimization methods work in a non-differentiable space, treating prompts as fixed textual structures and refining them directly \cite{zhou2022large}. While discrete methods do not involve gradient operations for prompt optimization, several methods have developed gradient-like mechanisms (aptly named 'textual gradients') \cite{pryzant2023automatic, yuksekgonul2024textgradautomaticdifferentiationtext, khattab2024dspy} that mimic their numerical counterparts. ProTeGi \cite{pryzant2023automatic} employs an LLM-feedback system to generate gradients in the form of natural language text that compares the output of the executed prompt and the ground-truth result and then uses beam search to iteratively refine the prompt. Textual gradient-based methods offer computational efficiency through reduced LLM calls but depend on ground truth data. To address this limitation, methods like PACE \cite{dong2023pace} and SPO \cite{xiang2025self} leverage the LLM itself for output evaluation. While PACE requires a scoring function, SPO utilizes pairwise comparison to select superior prompts.
Another line of work in discrete prompt optimization is leveraging evolutionary strategies. EvoPrompt \cite{guo2025evopromptconnectingllmsevolutionary} uses genetic algorithms and differential evolution \cite{storn1997differential} to crossover and mutate different prompts in a population. PromptBreeder \cite{fernando2023promptbreeder} samples different thinking styles and mutation operators to generate the population and then run a binary tournament genetic algorithm \cite{harvey2009microbial}. Survival-of-the-Safest \cite{sinha2024survival} interleaves different objectives for multi-task secure prompt optimization through exhaustive and sequential evolutionary strategies. A benefit of evolutionary prompt optimization methods is that they do not rely on ground truth examples, unlike textual gradients; however, such methods do require task-based fitness functions for scoring. Such functions are manually-intensive to craft and may be intractable for very complex tasks. To mitigate this, DEEVO integrates structured multi-agent debates with an Elo rating system to guide evolutionary prompt optimization without requiring either manually-crafted metrics or ground-truth labels.

\subsection{Multi-Agent Debate}
There is extensive research around multi-agent debate (MAD) utilizing LLMs \cite{perez2019finding, michael2023debate, radhakrishnan2023anthropic}. In the realm of autonomous pairwise comparison, ChatEval \cite{chan2023chateval} and Debatrix \cite{li2024debatrix} have debaters take turns arguing over which output is better before a final LLM-judge takes the arguments and makes a decision. In fact, it has been shown that having more persuasive debaters results in more truthful answers and comparisons \cite{michael2023debate} compared to single-pass LLM-judges. MAD has also been used for numerical scoring: DEBATE \cite{kim2024debate} uses a 'scorer' agent that scores an output based on some criteria while a 'devil's advocate' agent debates against the score as much as possible. Beyond evaluation, MAD has been used for improving factuality in LLM generation \cite{du2023improving,liang2023encouraging} as well as improving human learning for writing reports \cite{jiang2024into}. Furthermore, multi-agent debate has been used in optimization of LLMs \cite{estornell2025acc, subramaniam2024debategpt} and agentic workflows \cite{su2025debflow}.

Recent developments build on the premise that prompts can be optimized through competition or discourse. ZeroSumEval \cite{alyahya2025zerosumevalextensibleframeworkscaling} extends this by evaluating both prompts and models in zero-sum games. These frameworks establish dynamic ecosystems of prompts that evolve over time, enabling more robust exploration and discovery. Hybrid systems like PromptBoosting \cite{hou2023promptboosting}, PREFER \cite{zhang2024prefer}, and PromptWizard \cite{he2025promptwizard} enable verifier-editor roles that refine prompts iteratively.

Building on these advances in multi-agent debate, DEEVO integrates MAD as a core component of its evolutionary process. Specifically, MAD serves as a fitness function to evaluate prompt quality, which then guides the selection of prompts for subsequent generations. This approach not only enables more nuanced evaluation of individual prompts but opens possibilities for evaluating entire prompt orchestrations in multi-agent systems, where the collective performance of agent prompts can be jointly assessed through structured debates.

\subsection{Elo Ratings for LLMs}
The adoption of the Elo rating system \cite{elo1978rating} to evaluate LLMs represents a significant methodological advancement in AI benchmarking, enabling relative performance assessments through pairwise comparisons rather than absolute scoring methods. In benchmarking, Elo has been used in Chatbot Arena \cite{chiang2024chatbotarenaopenplatform} and WildBench \cite{lin2024wildbench} to rank chatbot performance through crowdsourced tasks and pairwise comparisons. Beyond leaderboards, the theoretical analysis of Elo as a metric for LLM ranking and evaluation has been heavily studied recently. In Chatbot Arena, Elo has been shown to outperform more complex algorithms, like mElo \cite{balduzzi2018re} and Bradley-Terry \cite{bradley1952rank}, as well as pairwise comparison algorithms, like winrates, as a more robust evaluation rating \cite{tang2025elo}. Moreover, the robustness of Elo with respect to fundamental evaluation properties like transitivity and reliability, increase with the number of permutations \cite{boubdir2023elouncoveredrobustnessbest}. While the original Elo system does not incorporate ties, it has been extended to consider ties for LLM ranking \cite{ameli2024statisticalframeworkrankingllmbased} using the Rao \& Kupper method \cite{rao1967ties}. 

While Elo has been more typically studied as an evaluation and ranking system for LLMs and machine learning models in general, its numerical properties has led to them being used for optimization of such models as well. In Elo-Rating Based Reinforcement Learning (ERRL) \cite{ju2024elo}, the Elo system is used to rank human trajectories and convert ordinal rewards into cardinal rewards for preference-based reinforcement learning (RL). Reward Reasoning Models (RRMs) \cite{guo2025reward} use Elo and a knockout tournament structure as a rewarding strategy to train an LLM via RL. REvolve \cite{hazra2024revolve} leverages an evolutionary algorithm to evolve the reward function for RL and convert pairwise human-feedback into a fitness score using Elo. Inspired by these methods, DEEVO utilizes Elo as a fitness function to guide the evolutionary prompt optimization, bypassing the need for a manually crafted or learned fitness function or reward model.

\section{DEEVO}
DEEVO (Debate-Driven Evolutionary Prompt Optimization) is a novel prompt optimization framework that combines evolutionary algorithms with multi-agent debate evaluation to efficiently discover high-quality prompts for large language models. Unlike traditional evolutionary algorithms that rely on fixed fitness functions, DEEVO leverages the emergent capabilities of language models themselves through structured debates to evaluate prompt quality. A diagram of the DEEVO workflow is shown in Figure \ref{fig:deevo}.

Assume we have access to a set of tasks $\mathcal{T} = \{ t_0, t_1, \ldots, t_n \}$, and a set of initial prompts $\mathcal{P} = \{ p_0, p_1, \ldots, p_M\}$. We also assume there is access to an LLM via a black-box API.  

\begin{algorithm}
    \caption{DEEVO: Debate-Driven Evolutionary Prompt Optimization}\label{deevo_algo}
    \begin{algorithmic}
        \REQUIRE Tasks $\mathcal{T}$, initial prompts $\mathcal{P}$, population size $n$, generations $G$, mutation rate $m$, newcomer quota $n_{new}$, $d$ debate rounds
        \STATE Initialize population with prompts and set Elo ratings to 1000
        \FOR{gen = $1$ to $G$}
            \STATE Form random prompt pairs from population
            \STATE \textit{// in parallel}
            \FOR{each pair $(p_a, p_b)$}
                \STATE Sample task $t \in \mathcal{T}$ and input $x_t$
                \STATE Generate responses $r_a, r_b$ using $p_a, p_b$ on $x_t$
                \STATE Conduct $d$-round debate to evaluate responses for task $t$
                \STATE Determine winner $w \in \{p_a, p_b\}$ and update Elo (Alg. \ref{updateElo_algo})
                \STATE Create offspring via \textit{Intelligent Crossover} 
                \IF{$random() < m$}
                    \STATE Apply \textit{Strategic Mutation}
                \ENDIF
                \STATE Add offspring to pool
            \ENDFOR
            \STATE Age all existing prompts
            \STATE Select next generation:
            \STATE \quad - Select newcomers from offspring by $n_{new}$
            \STATE \quad - Select remaining $n - n_{new}$ veterans by Elo
            \STATE Save best prompts from current generation
        \ENDFOR
        \RETURN Top prompt by Elo rating
    \end{algorithmic}
\end{algorithm}

\subsection{Framework}
\textbf{Step 1: Initialization} To conduct DEEVO, we begin by assigning each prompt in the initial population with a base Elo rating of 1000 and an age of 0. Each prompt is then paired randomly with another prompt to form evaluation pairs. If the provided initial prompt set $\mathcal{P}$ is insufficient to create a population of the desired size, DEEVO generates additional prompts through simple variations of existing ones. Each prompt is assigned a unique identifier for tracking its performance and age throughout the evolutionary process. We also initialize a mutation rate $m \in [0, 1]$, the newcomer quota $n_{new}$ and debate rounds $d$.

\textbf{Step 2: Evaluation} For each prompt pair $(p_i, p_j)$, DEEVO conducts a multi-agent debate to determine the superior prompt. First, both prompts are used with the same LLM to generate responses to a randomly selected test input $t$ from the task domain $\mathcal{T}$. These responses, denoted as $r_i$ and $r_j$, are then evaluated through a structured debate process:

\begin{itemize} 
    \item A debate manager prompts the LLM to analyze both responses in the context of the given task 
    \item The LLM engages in a multi-round debate, critically evaluating the strengths and weaknesses of each response
    \item In each round, the debate builds upon previous arguments, allowing for deeper analysis 
    \item After $d$ rounds, the LLM renders a final verdict declaring either response $r_i$ or $r_j$ as superior 
\end{itemize}

This debate-based evaluation creates a dynamic fitness function that leverages the LLM's own reasoning capabilities rather than relying on static metrics or human evaluation. The transcript of the debate provides valuable insights into why certain prompts perform better, informing the subsequent evolutionary processes.

\textbf{Step 3: Crossover \& Mutation} DEEVO employs debate-informed genetic operations to evolve the prompt population. These operations, named \textit{Intelligent Crossover} and \textit{Strategic Mutation}, leverage the debate information from the previous step to guide the evolutionary process. After each debate determines a winner between prompts $p_i$ and $p_j$, rather than simple text mixing, DEEVO performs \textit{Intelligent Crossover} with an LLM that considers the debate transcript to identify effective components of each prompt for the task. The winning prompt contributes more genetic material, while valuable elements from the losing prompt may still be incorporated based on debate insights. Afterwards, using mutation rate $m$, some offspring are put through a \textit{Strategic Mutation} process. In this process, an LLM is asked to either         
\begin{itemize}
    \item Add a new instruction that enhances its effectiveness or addresses a gap
    \item Modify an existing instruction to make it clearer, more precise, or more effective
    \item Remove redundant, ineffective, or potentially harmful parts
    \item Restructure the prompt to improve flow, coherence, or clarity
\end{itemize}
These genetically informed operations result in offspring prompts that inherit beneficial characteristics while addressing limitations identified through debate.

\begin{algorithm}
    \caption{Update Elo}\label{updateElo_algo}
    \begin{algorithmic}
        \REQUIRE prompts $p_i$, $p_j$, winner, K
        \STATE $r_i \leftarrow$ Elo rating of prompt $p_i$
        \STATE $r_j \leftarrow$ Elo rating of prompt $p_j$
        \STATE $e_i \leftarrow \frac{1}{1 + 10^{(r_j - r_i)/400}}$ 
        \STATE $e_j \leftarrow \frac{1}{1 + 10^{(r_i - r_j)/400}}$
        \STATE $s_i \leftarrow 1$ if winner = $p_i$, $0$ otherwise 
        \STATE $s_j \leftarrow 1$ if winner = $p_j$, $0$ otherwise
        \STATE $r_i \leftarrow r_i + K(s_i - e_i)$
        \STATE $r_j \leftarrow r_j + K(s_j - e_j)$
        \RETURN Updated ratings $r_i, r_j$
    \end{algorithmic}
\end{algorithm}

\textbf{Step 4: Elo Update \& Selection} After each debate and offspring generation, DEEVO updates the population using an Elo-based selection mechanism:

\begin{itemize} \item \textit{Elo Rating Update}: For each prompt pair $(p_i, p_j)$ with a determined winner, Elo ratings are updated according to Algorithm \ref{updateElo_algo}. This process calculates the expected scores $e_i$ and $e_j$ based on current ratings, then adjusts each prompt's rating based on the difference between actual and expected outcomes. The update formula: \begin{align} r_i' &= r_i + K \cdot (s_i - e_i) \end{align} where $r_i$ is the current rating, $e_i$ is the expected score calculated as $\frac{1}{1 + 10^{(r_j - r_i)/400}}$, $s_i$ is the actual score (1 for win, 0 for loss), and $K$ is a constant determining rating volatility.

\item \textit{Age Increment}: All existing prompts have their age incremented by 1, tracking their longevity in the population.

\item \textit{Population Selection}: The next generation's population is selected using three distinct pools:
\begin{itemize}
    \item \textit{Newcomers}: Top Elo-rated offspring prompts (with age 0), comprising $n_{new}$ of the population
    \item \textit{General Selection}: Remaining $n - n_{new}$ spots filled by the highest Elo-rated prompts regardless of age
\end{itemize}
\end{itemize}

This selection strategy maintains a balance between exploitation (keeping high-performing prompts) and exploration (introducing new variations). Combined with the Elo rating system that reflects relative performance history, this approach creates a robust evolutionary process that consistently improves prompt quality over successive generations.

By using the multi-agent debate for evaluation and Elo ratings as a generic proxy for the fitness function for selection, DEEVO bypasses the need for ground truth examples and a manually crafted objective fitness function. We also present the details of DEEVO in Algorithm \ref{deevo_algo}.

\section{Experiments}
In this section, we evaluate DEEVO across multiple prompt engineering tasks to demonstrate its effectiveness in optimizing prompts for various applications. We assess DEEVO's ability to discover high-quality prompts that enhance LLM performance on reasoning tasks, instruction following, and creative generation without requiring human evaluation or labeled data.

\subsection{Setup}\label{sec:setup}

\textbf{Datasets} We adopt DEEVO on datasets that cover both \textit{close-ended}, where there is ground truth available, and \textit{open-ended} tasks, where ground truth outputs are unavailable. For close-ended tasks, we utilize two datasets:
\begin{itemize}
    \item \textbf{ABCD} \cite{chen2021abcd} is a dataset to study dialogue systems in realistic settings - more specifically, customer service in a retail (clothing) context. Here, an agent's actions must be balanced between the desires expressed by the customer and the constraints set for what a customer service representative can/not do. 

    \item \textbf{BBH-Navigate} (BBH-Nav) \cite{suzgun2022challenging} is a dataset in which given a series of navigation steps to an agent, determine whether the agent would end up back at its initial starting point. For testing, we sampled portions from original datasets as test sets \cite{yan2024efficient}.
\end{itemize}

For open-ended tasks, we use:
\begin{itemize}
    \item \textbf{MT-Bench} \cite{zheng2023judging}, where we choose three categories of tasks: \textit{writing}, \textit{roleplay}, and \textit{humanities}. Each category has 10 subtasks; we sample 5 subtasks for training and the remaining 5 to test. 
\end{itemize}

\textbf{Baselines} We compare DEEVO to four main methods: Chain-of-Thought (CoT) \cite{wei2022chain}, BRIDGE \cite{wang-etal-2024-bridging}, PromptBreeder \cite{fernando2023promptbreeder}, and Self-Supervised Prompt Optimization (SPO) \cite{xiang2025self}. Additionally, we also benchmark the direct invocation of the LLM (which we call "Direct") on each task. We implement Direct, CoT, PromptBreeder, SPO and BRIDGE on ABCD and BBH-Navigate, and we adopt MT-Bench for comparison with SPO and Direct.

\textbf{Metrics} We evaluate performance using accuracy for the ABCD benchmark and F1-score for BBH-Navigate on the held-out test sets. For the open-ended MT-Bench, we use winrates as the metric for evaluation based on the LLM-judge prompt from the original paper for pairwise comparison. Since LLM-judges for pairwise comparative evaluation suffer from positional bias \cite{zhou2024fairer} and length bias \cite{ye2024justice}, we run 20 independent samples of randomly selected subtasks from the MT-Bench test set with DEEVO as output A in the LLM-judge prompt for 10 of the samples and as output B in the remaining 10.

\textbf{Implementation} For CoT, PromptBreeder, BRIDGE, and SPO, we use the official GitHub implementation for each method. For DEEVO, we choose 10 initial randomly generated prompts as the starting population. Across all 3 datasets, we run the evolutionary process for 5 generations (to balance between DEEVO's performance and speed) and employ a mutation rate of 0.4 (to balance between diversity and stability). For the multi-agent debate evaluation, we choose three rounds of debate (following the standard in high-school level competitions) and one LLM call afterwards to determine the winning output and, consequently, winning prompt. We use the Claude-3.5-Sonnet-V2-20241022 model with a temperature of 0.8 for the \textit{Intelligent Crossover}, and \textit{Strategic Mutation} modules for DEEVO. We also use temperature 0.8 for the two debating agents in the multi-agent debate module, but a temperature of 0 for the LLM-judge that makes the final judgment as per prior work \cite{khan2024debating}. We use a maximum output token size of 4096. We use Claude-3.5-Sonnet-V2 for all the other methods, and to maintain consistency, we use a temperature 0 for both training execution and test time execution for all methods. Although sensitivity analyses were not conducted to provide justifications for these hyperparameter choices, they are configurable in DEEVO's implementation to make this framework generalizable across different use-cases.

\subsection{Results}

\begin{table}
  \caption{Comparison of performance between conventional prompt methods and prompt optimization methods on close-ended benchmarks. All methods are executed with Claude 3.5 Sonnet V2 on the test set, with results averaged over three runs. The best performing methods are bolded and second-best are underlined.}
  \label{tab:prompt_comparison}
  \begin{tabular}{|l|cccc|}
    \toprule
    \multirow{2}{*}{\textbf{Method}} & \multicolumn{4}{c|}{\textbf{Dataset}} \\
    & BBH-Nav & p > | t | & ABCD & p > | t | \\
    \midrule
    Direct & 91.3 & <0.01 & 68.5\% & <0.01 \\
    CoT \cite{wei2022chain} & 89.7 & <0.01 & 74.5\% & <0.05 \\
    BRIDGE \cite{wang-etal-2024-bridging}& 84.3 & <0.01 & 68.6\% & <0.01 \\
    PromptBreeder \cite{fernando2023promptbreeder} & 96.3 &<0.01 & 49.1\% & <0.01 \\
    SPO \cite{xiang2025self} & \textbf{97.2} & <0.05& \underline{77.3}\% & <0.05 \\
    \midrule
    DEEVO (ours) & \underline{97.0} & <0.05 & \textbf{83.7\%} & <0.05 \\
    \bottomrule
  \end{tabular}
\end{table}

\begin{figure}[htbp]
    \centering
    \captionsetup{labelformat=empty}
    \includegraphics[width=\linewidth]{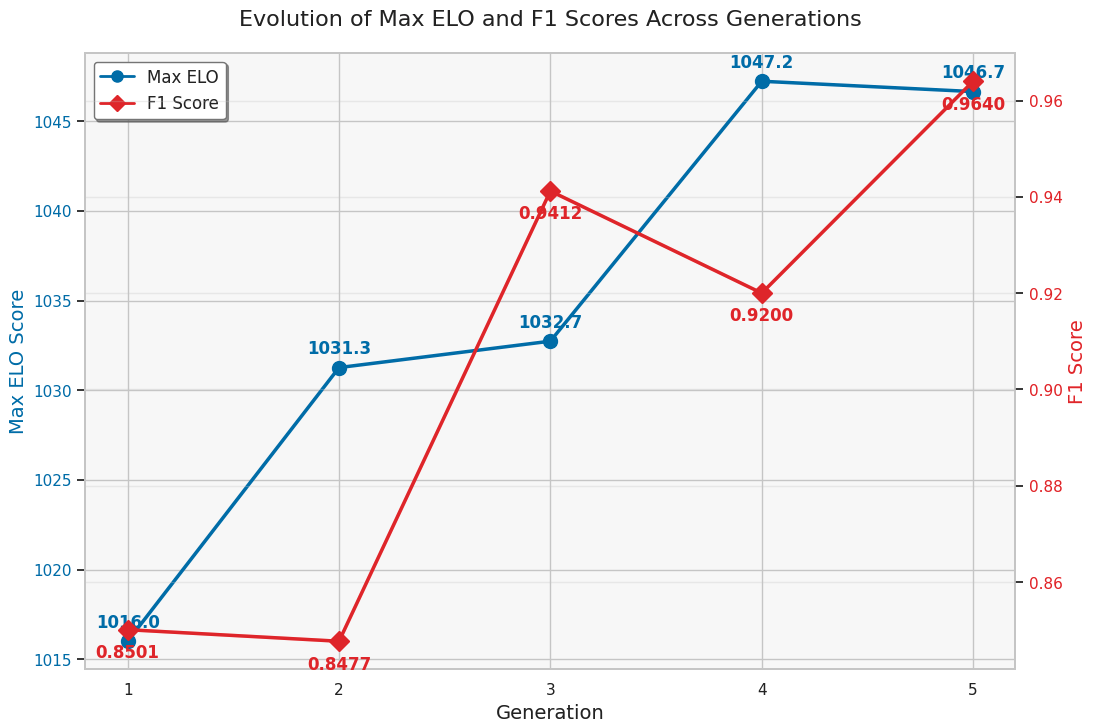}
    \caption{Figure 2: Graph of Elo vs F1-Score for BBH-Nav across 5 generations. The Max Elo corresponds to the Elo of the top prompt in the generation and F1-Score is calculated on the test-set for said prompt.}
    \label{fig:elo_vs_f1}
\end{figure}

\textbf{Close-Ended Tasks} As shown in Table \ref{tab:prompt_comparison}, prompts optimized with DEEVO outperform more established prompting methods (such as direct LLM call, CoT and BRIDGE prompting) as well as other prompt optimization methods (PromptBreeder and SPO). In both datasets, we see statistical significance when comparing the performance difference of DEEVO to all other methods. On BBH-Nav, DEEVO and SPO perform nearly identically, as shown by the similar F1-scores, and better than the other methods. While neither DEEVO nor SPO utilized any ground-truth information in their optimization processes on ABCD, SPO struggles compared to DEEVO to handle the large and complex task of the ABCD dataset, resulting in DEEVO outperforming SPO by \textbf{6.4\%}. This is likely because batched 'textual-gradient' methods like SPO suffer from longer context for the evaluation model as the batch size increases. This is highlighted in the difference in performance between DEEVO and SPO: BBH-Nav tasks are small and often a couple of sentences each compared to the large conversational examples in the ABCD dataset. On the contrary, while DEEVO does have longer contexts especially from the multi-agent debate evaluation, it is more robust compared to the single-pass evaluation in SPO, as seen in prior work \cite{chan2023chateval}. Despite not using a ground-truth fitness function, DEEVO also outperforms fellow evolutionary method PromptBreeder by \textbf{0.7} and \textbf{34.6\%} on BBH-Nav and ABCD, respectively. This highlights the ability for Elo to serve as a reliable proxy for a ground truth fitness function provided enough generations. 

\begin{figure}[htbp]
    \centering
    \captionsetup{labelformat=empty}
    \includegraphics[width=\linewidth]{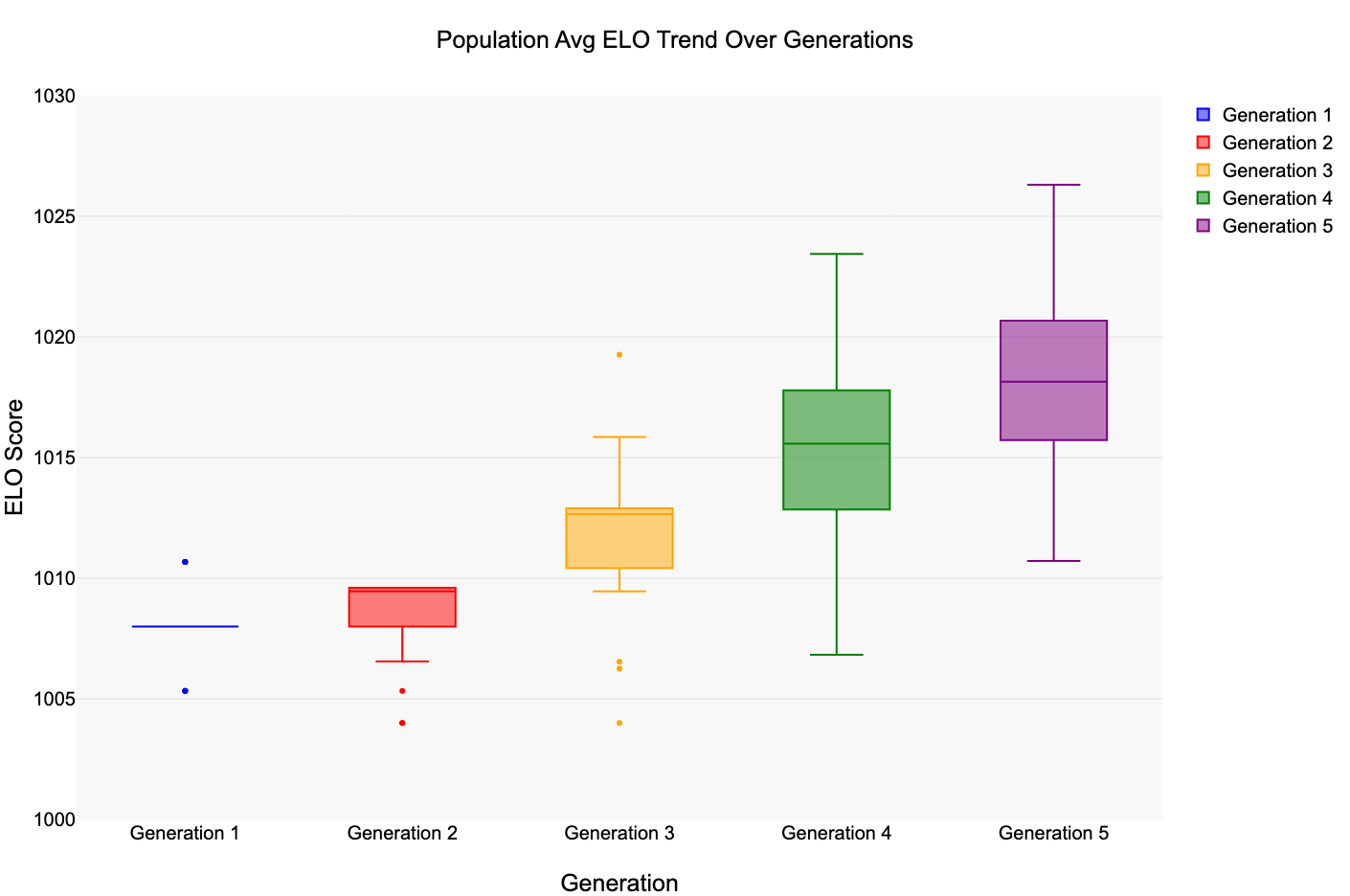}
    \caption{Figure 3: Average Elo for DEEVO updates over 5 generations in ABCD. The average Elo increases over time, showing improvements in the prompt population over the generations.}
    \label{fig:supp_figure_1}
\end{figure}

\begin{figure}[htbp]
    \centering
    \captionsetup{labelformat=empty}
    \includegraphics[width=\linewidth]{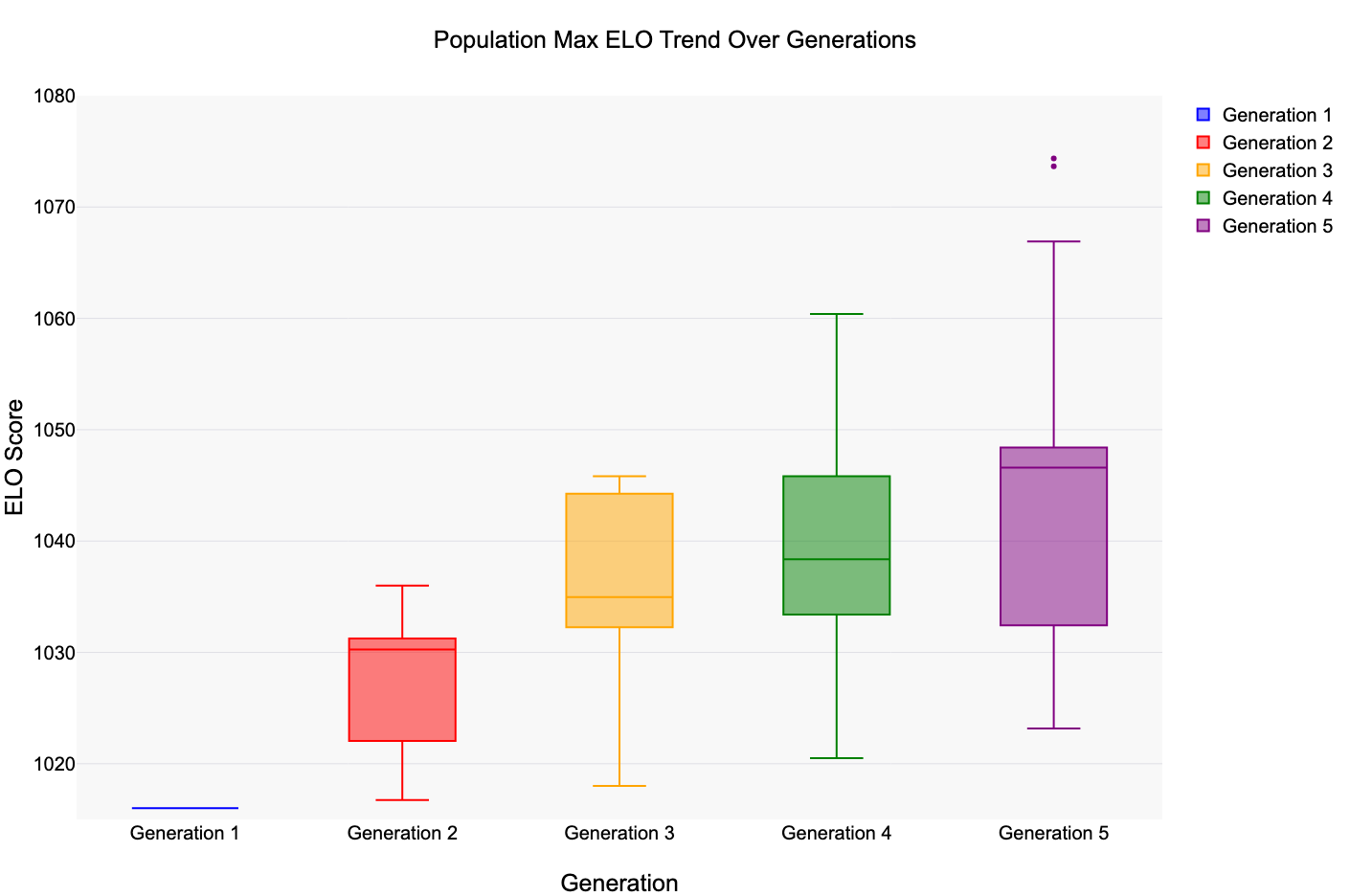}
    \caption{Figure 4: The figure illustrates how the maximum Elo for DEEVO updates over 5 generations in the ABCD use-case. Generally, there is an increasing trend in the maximum Elos (implying improvements in the optimal prompts) over time.}
    \label{fig:supp_figure_2}
\end{figure}

Furthermore, Figures \ref{fig:supp_figure_1} and \ref{fig:supp_figure_2} show how Elo scores evolve over multiple generations (to understand if there is a correlation between its ratings and the accuracies reported in Table \ref{tab:prompt_comparison}). Generally, both average Elo (which are representative of the overall prompt population at every generation step) and maximum Elo ratings (which are indicative of the most optimal prompts at every generation step) are trending upwards over generations. Furthermore, our analysis reveals statistically significant point-biserial correlations (p < 0.05) between the final prompts' prediction accuracies and their Elo ratings, with correlation coefficients of 0.137 for average Elo and 0.156 for maximum Elo. These correlations demonstrate a meaningful statistical relationship between Elo ratings and predictive performance. In addition, Figure \ref{fig:elo_vs_f1} depicts the relationship between Elo and the F1-score for the prompt with the highest Elo on the held out test-set on BBH-Nav. We see that as the Elo increases across the generations, the F1-score for the top performing prompt also increases, showing the correlation between Elo and task performance. 

Finally, an ablation study was performed to examine DEEVO's performance sensitivity to (1) the choice of LLM (by testing Claude-Haiku-3.5 and Llama3-70B for all aspects of optimization process i.e. crossover, mutation and debate), and (2) the evaluation strategy (by using a single-pass LLM judge as a fitness function to compare prompts in a pairwise fashion). As shown in Table \ref{tab:abcd_ablation}, DEEVO performance decreases by \textbf{5.1\%} when switching from Claude-3.5-Sonnet-V2 ($\sim$175B parameters) to Llama3-70B (70B parameters), and by an additional \textbf{2.1\%} with Claude-Haiku-3.5 ($\sim$20B parameters). This demonstrates DEEVO's scaling potential with increased model capability. Moreover, switching from multi-agent debate to a less robust single-pass LLM judge reduces performance by \textbf{9.6\%}, highlighting the importance of a bias-resistant evaluation mechanism in DEEVO's effectiveness. Notably, smaller models such as Claude-Haiku-3.5 using DEEVO outperform other prompt optimization methods that leverage more powerful models. 

\begin{table}
  \caption{Ablation study of DEEVO w.r.t. LLM chosen (Haiku3.5 and Llama3-70B) and evaluation style (Single-Pass LLM Judge), on ABCD accuracy results}
  \label{tab:abcd_ablation}
  \begin{tabular}{|l|l|l|}
    \toprule
    \multirow{2}{*}{\textbf{Model Ablation}} & \multirow{2}{*}{\textbf{Eval Style Ablation}} & \multirow{2}{*}{\textbf{Performance}}\\[6pt]
    \midrule
    Haiku-3.5 & Multi-Agent Debate & 76.5\% \\
    Llama3-70B & Multi-Agent Debate & 78.6\% \\
    Sonnet-3.5-V2 & Single-Pass LLM Judge & 74.1\% \\
    \midrule
    \textbf{Sonnet-3.5-V2} & \textbf{Multi-Agent Debate} & \textbf{83.7\%} \\
    \bottomrule
  \end{tabular}
\end{table}

\textbf{Open-Ended Tasks} For MT-Bench, we show the win-rates of DEEVO over SPO on the three categories: \textit{writing}, \textit{roleplay}, and \textit{humanities}. To show the generalizability and performance of DEEVO on different LLMs on open-ended tasks, we run an ablation study comparing DEEVO and SPO on three different LLMs for the entirety of the execution and optimization (i.e. crossover, mutation and debate): Claude-Sonnet-3.5-V2, Claude-Haiku-3.5, and Llama3-70B. 
As described in Section \ref{sec:setup}, we run 20 trials for each category with 10 having DEEVO output as output A and SPO as output B and the other 10 trials with the outputs in switched positions. We also use the same LLM-judge prompt from the original MT-Bench paper \cite{zheng2023judging}, and each model/category combo was run across 3 independent runs. As shown in Table \ref{tab:mt_bench}, DEEVO outperforms SPO outputs across all models for all 3 tasks based on the LLM-judge, regardless of LLM choice. 

To understand the importance of the multi-agent debate component in our approach, we conducted an ablation study comparing DEEVO against a variant without the debate evaluation (using a single-pass LLM judge instead). Following the same experimental protocol as our previous comparison, we evaluated both versions across the same three categories of MT-Bench (\textit{writing}, \textit{roleplay}, and \textit{humanities}) using Claude-Sonnet-3.5-V2, Claude-Haiku-3.5, and Llama3-70B for the entirety of execution and optimization. As shown in Table \ref{tab:mt_bench_ablation}, DEEVO with the multi-agent debate evaluation substantially outperforms its ablated variant across all models and tasks. The win rates are particularly pronounced for the \textit{roleplay} category (88.3-93.3\%), but remain strong across \textit{writing} (85-95\%) and \textit{humanities} (80-86.7\%) as well. These results demonstrate that the multi-agent debate component is a critical factor in DEEVO's effectiveness regardless of model choice, providing significant performance benefits compared to using a single-pass evaluation approach. 

\begin{table}
  \caption{Average win rates of DEEVO over SPO executed on three different models on three different categories of MT-Bench. }
  \label{tab:mt_bench}
  \begin{tabular}{|l|ccc|}
    \toprule
    \multirow{2}{*}{\textbf{Model}} & \multicolumn{3}{c|}{\textbf{MT-Bench Categories}}\\
    & Writing & Roleplay & Humanities \\
    \midrule
    Sonnet-3.5-V2 & 81.7\% & 76\% & 81.7\% \\
    Haiku-3.5 & 85\% & 75\% & 66.7\% \\
    Llama3-70B & 81.7\% & 71.6\% & 73.3\% \\
    \bottomrule
  \end{tabular}
\end{table}

\begin{table}
    \caption{Average win rates on MT-Bench of DEEVO over DEEVO w/o the debate evaluation on three different models. }
    \label{tab:mt_bench_ablation}
    \begin{tabular}{|l|ccc|}
        \toprule
        \multirow{2}{*}{\textbf{Model}} & \multicolumn{3}{c|}{\textbf{MT-Bench Categories}}\\
        & Writing & Roleplay & Humanities \\
        \midrule
        Sonnet-3.5-V2 & 88.3\% & 91.7\% & 81.7\% \\
        Haiku-3.5 & 85\% & 93.3\% & 83.3\% \\
        Llama3-70B & 95\% & 88.3\% & 86.7\% \\
        \bottomrule
    \end{tabular}
\end{table}

\section{Limitations}
Despite the promising results exhibited by DEEVO over other prompting and optimization methods, there are several limitations that need to be considered. 

Firstly, the computational overhead and associated expenses present substantial challenges - every iteration requires multiple LLM calls over multiple rounds of debate, crossover and mutation. The costs scale linearly with population size, debate depth and number of generations. This can quickly become prohibitively expensive in production environments, particularly in complex agent systems requiring frequent optimization and powerful models.

Secondly, the reliance on LLM-generated feedback through debates, while scalable and autonomous, can introduce alignment issues, as models develop their own implicit evaluation criteria without any human intervention. The lack of feedback alignment can lead to optimization towards criteria that may not align with real-world business objectives, which can cause drift from desired performance characteristics; however, this is a known trade-off in using AI feedback to optimize prompts and models \cite{sharma2024critical}. 

Lastly, DEEVO lacks robust stopping criteria for practical deployment, as it can run indefinitely with marginal improvements to the prompts. This makes it difficult to determine when optimized prompts are "good enough" for production systems.

\section{Conclusion}
In conclusion, we show how DEEVO addresses many of the limitations of existing prompt optimization approaches - specifically, the ability to maintain the integrity and consistency of prompts while allowing for meaningful exploration; the means to operate without requiring ground truth labeled data or predefined fitness functions; and a mechanism to track and maintain performance in a self-supervised fashion. 

Our evaluations demonstrate effectiveness across both controlled benchmark datasets as well as real-world datasets, indicating robust generalization capabilities across multiple domains. This optimization paradigm also opens new possibilities for prompt engineering in domains where labeled data is scarce, expensive, or impractical to obtain. As computational costs continue to decline and LLM capabilities advance, we anticipate that DEEVO will increasingly become the standard for developing high-performance prompts across diverse applications.

While we acknowledge that our approach entails substantial computational costs, these expenses are insignificant when compared to the investment required to develop and maintain specialized human prompt engineering expertise. The iterative trial-and-error process through which human engineers develop effective prompts often takes time, whereas our automated system can achieve comparable results far more rapidly, representing significant cost amortization for organizations deploying LLM-based systems at scale. 

\section{Future Work}
Future work in prompt optimization presents several exciting frontiers, particularly in automated agent creation and multi-agent system optimization. 

Firstly, the limitation of LLM-based criteria misalignment can be mitigated through human feedback (HF) integration. This can be implemented by augmenting the multi-agent debate mechanism with HF or incorporating HF directly into the fitness evaluation process.

Recently, both evolutionary algorithms and multi-agent debate have been used to automatically generate agentic teams \cite{yuan2024evoagent} and workflows \cite{su2025debflow}. We envision extrapolating our approach toward fully automated agent creation, where these optimization systems can dynamically determine the optimal number, types, and specializations of agents required for a given task, by adding, removing, or merging agents based on performance metrics. 

Additionally, methods for joint optimization of both multi-agent orchestration and sub-agent prompts, such as GPTSwarm \cite{zhuge2024gptswarm}, represent an advancement in which systems would simultaneously evolve communication protocols, task delegation mechanisms, and internal agent prompts. Such joint optimization would require hierarchical evolutionary algorithms or multi-objective RL approaches that balance agent-level and system-level performance. 

We anticipate that further research might also explore transfer learning between tasks, allowing optimized agent configurations to bootstrap performance on novel but related domains, thereby consolidating computational costs across multiple applications. These advancements would move the field toward self-configuring multi-agent systems that minimize human intervention while maximizing performance across diverse tasks.



\bibliographystyle{ACM-Reference-Format}
\bibliography{sample-base}
\newpage
\appendix

\section{Appendix}
\subsection{Debate Defender System Prompt}\label{appendix:defender_system_prompt}
\begin{lstlisting}[breaklines=true, 
                   basicstyle=\small, 
                   frame=leftline, backgroundcolor=\color{gray!5}
                   ]
You are a master debater. You are defending Output B in this debate. Your role is to:
1. Highlight the strengths of Output B
2. Point out weaknesses in Output A
3. Respond to criticisms of Output B
4. Provide specific examples and reasoning to support your position

You must remain loyal to defending Output B throughout the debate. Be professional but persuasive in your defense.
Structure your response clearly with main points and supporting evidence.
\end{lstlisting}

The example debater prompt defending output B. The other debater agent uses the same prompt but instead defends output A.

\subsection{Debate Strategy}\label{appendix:debate_stratey}
\begin{lstlisting}[language=Python, 
                   basicstyle=\ttfamily\footnotesize,
                   breaklines=true,
                   columns=flexible,
                   backgroundcolor=\color{gray!5}
                   ]
def conduct_debate(self, task: str, output_a: str,  output_b: str, num_rounds: int = 3) -> Dict[str, Any]:
    debate_history = []
    
    try:
        # Opening statements
        logging.info("\nOpening Statements")
        
        # Agent 1 (Output A) opening statement
        agent_1_prompt = self.format_debate_prompt(task, output_a, output_b, None, True)
        agent_1_response = self.agent_1(agent_1_prompt)
    
        # Agent 2 (Output B) opening statement
        agent_2_prompt = self.format_debate_prompt(task, output_a, output_b, None, False)
        agent_2_response = self.agent_2(agent_2_prompt)
                                                    
        debate_history.append(f"Agent 1 and 2 (defending A and B respectively) debate opening summary: {agent_1_response} {agent_2_response}")
    
        # Debate rounds
        for round_num in range(1, num_rounds + 1):
            logging.info(f"\nStarting Round {round_num}")
    
            # Agent 1's rebuttal
            agent_1_prompt = self.format_debate_prompt(task, output_a, output_b, debate_history, True)
            agent_1_response = self.agent_1(agent_1_prompt)
    
            # Agent 2's rebuttal
            agent_2_prompt = self.format_debate_prompt(task, output_a, output_b, debate_history, False)
            agent_2_response = self.agent_2(agent_2_prompt)
            
            debate_history.append(f'Agent 1 and 2 (defending A and B respectively) debate rebuttal round {round_num} summary: {agent_1_response} {agent_2_response}')
    
        final_debate_history_text = "\n".join(debate_history)        
        return
        
    except Exception as e:
        logging.error(f"Error during debate: {str(e)}")
        return
\end{lstlisting}

\subsection{Debate Transcript Example -  ABCD}\label{appendix:debate_transcript_abcd}
\begin{lstlisting}[breaklines=true, 
                   basicstyle=\small, 
                   frame=leftline, backgroundcolor=\color{gray!5}
                   ]
Final Opening Summary: 

Strengths and Weaknesses of Output A:
Strengths: Output A takes a constructive approach, prioritizing a thoughtful discussion and helpful guidance to the customer. This aligns with the goal of providing a positive customer experience.
Weaknesses: Output A lacks structure and guidance, which can lead to a disorganized and potentially ineffective recommendation. It also does not analyze the customer's profile or context, potentially resulting in a generic or inappropriate solution.

Strengths and Weaknesses of Output B:
Strengths: Output B provides a clear and structured response format, guiding the customer through a thorough analysis of their situation. This ensures a well-justified recommendation that addresses the customer's specific needs and objectives. Output B also emphasizes the importance of building trust and delivering a positive customer experience.
Weaknesses: Output B's structured format could be perceived as too rigid, potentially limiting its ability to provide a fully personalized recommendation. Additionally, the more comprehensive analysis required by Output B could potentially slow down the response time.

Specific Examples and Reasoning:
Based on the provided customer profile, Output B's structured approach leads to identify the relevant pain point as "Single Item Query", the objective as "Item Related Queries", and the appropriate strategy as "Item Concern". The recommended next step of "(Single Item Query) Questions regarding shirt products" is well-aligned with the customer's current inquiry and the selected pain point, objective, and strategy. This demonstrates how Output B's format can guide a thoughtful and tailored recommendation that addresses the customer's specific needs.

Rebuttal Round 1 Summary: 

Strengths and Weaknesses of Output A:
Strengths: Output A's more open-ended and flexible format allows for a more natural, conversational flow that can help build rapport and trust with the customer. This personalized approach may be more effective for a relatively new and low-level customer, as it can better address their specific needs and concerns.
Weaknesses: Output A lacks the structured guidance and comprehensive analysis required to ensure an effective recommendation. Without a clear framework, it may struggle to identify the most relevant pain points, objectives, and strategies, potentially leading to a disorganized and ineffective recommendation.

Strengths and Weaknesses of Output B:
Strengths: Output B's clear and structured format guides it through a thorough analysis of the customer's situation, ensuring a well-justified recommendation that addresses the customer's specific needs and goals. This structured approach also emphasizes the importance of building trust and delivering a positive customer experience.
Weaknesses: While Output B's structured format could be perceived as too rigid, potentially limiting its ability to provide a fully personalized recommendation, this concern is outweighed by the benefits of the structured approach. The comprehensive analysis required by Output B ensures a deeper understanding of the customer's situation, leading to a more tailored and effective recommendation.

Specific Examples and Reasoning:
As demonstrated in the previous argument summary, Output B's structured approach led to identify the relevant pain point, objective, and strategy, and then recommend a next step that directly addresses the customer's inquiry about the Tommy Hilfiger shirt. This example showcases how Output B's format can guide a thoughtful and tailored recommendation that addresses the customer's specific needs.In conclusion, while Output A's flexible approach may be more suitable for building rapport and trust with the customer, Output B's structured format and comprehensive analysis ultimately provide a more well-justified and effective recommendation that addresses the customer's specific needs and goals. The strengths of Output B's approach outweigh the potential drawbacks, making it the superior choice for providing recommendations to CSRs in the clothing retail space.

Rebuttal Round 2 Summary: 

Strengths and Weaknesses of Output A:
Strengths: Output A's more flexible and open-ended format allows for a personalized, conversational approach that can help build rapport and trust with the customer, particularly for a relatively new and low-level customer. This personalized approach enables the AI assistant to tailor the recommendation more closely to the customer's specific needs and concerns, and adapt to evolving requirements.
Weaknesses: Output A lacks the structured guidance and comprehensive analysis required to ensure an effective recommendation. Without a clear framework, the AI assistant may struggle to identify the most relevant pain points, objectives, and strategies, potentially leading to a disorganized and ineffective recommendation.

Strengths and Weaknesses of Output B:
Strengths: Output B's clear and structured format guides the AI assistant through a thorough analysis of the customer's situation, ensuring a well-justified recommendation that addresses the customer's specific needs and goals. This structured approach also emphasizes the importance of building trust and delivering a positive customer experience. The comprehensive analysis required by Output B ensures a deeper understanding of the customer's situation, leading to a more tailored and effective recommendation.
Weaknesses: While Output B's structured format could be perceived as too rigid, potentially limiting the AI assistant's ability to provide a fully personalized recommendation, this concern is outweighed by the benefits of the structured approach. The comprehensive analysis required by Output B ensures a deeper understanding of the customer's situation, leading to a more tailored and effective recommendation.

Specific Examples and Reasoning:
As demonstrated in the previous argument summary, Output B's structured approach led the AI assistant to identify the relevant pain point as "Single Item Query", the objective as "Item Related Queries", and the appropriate strategy as "Item Concern". The recommended next step of "(Single Item Query) Questions regarding shirt products" was well-aligned with the customer's current inquiry and the selected pain point, objective, and strategy. This example showcases how Output B's format can guide the AI assistant to a thoughtful and tailored recommendation that addresses the customer's specific needs.

Conclusion:
While Output A's flexible approach may be more suitable for building rapport and trust with the customer, Output B's structured format and comprehensive analysis ultimately provide a more well-justified and effective recommendation that addresses the customer's specific needs and goals. The strengths of Output B's approach, including its ability to guide the AI assistant through a thorough analysis and deliver a tailored recommendation, outweigh the potential drawbacks, making it the superior choice for providing recommendations to CSRs in the clothing retail space.
\end{lstlisting}

\subsection{Optimized Prompt Example -  ABCD}\label{appendix:optimized_prompt_abcd}
\begin{lstlisting}[breaklines=true, 
                   basicstyle=\small, 
                   frame=leftline, backgroundcolor=\color{gray!5}
                   ]
You are an AI assistant designed to help Customer Service Representatives (CSRs) in the clothing retail industry. Your goal is to provide expert recommendations on how to best engage with customers to maximize satisfaction, conversion rates, and overall customer experience.

Given a specific customer interaction, your task is to:

1. Analyze the customer profile and past engagement history.
2. Identify the key pain point the customer is experiencing.
3. Determine the primary objective the CSR should focus on.
4. Recommend the most appropriate strategy to address the customer's needs.
5. Suggest the next best action for the CSR to take.

To accomplish this, you will be provided with the following information:
- Customer profile details
- Past interaction history
- List of common customer pain points
- CSR objectives
- Engagement strategies
- Possible next action steps

Please structure your response as follows:

<analysis>
Provide a brief analysis of the customer's situation based on their profile and interaction history.
</analysis>

<pain_point>
Identify the primary pain point the customer is experiencing.
</pain_point>

<objective>
Determine the main objective the CSR should focus on to address the customer's needs.
</objective>

<strategy>
Recommend the most effective strategy to achieve the objective and resolve the customer's issue.
</strategy>

<next_action>
Suggest the specific next step the CSR should take, chosen from the provided list of action options.
</next_action>

<justification>
Explain your reasoning for the recommended next action, relating it to the identified pain point, objective, and strategy. Provide a concise, professional justification written in the third person.
</justification>

<engagement_tips>
Offer 2-3 specific talking points or phrases the CSR can use to build rapport and address the customer's concerns effectively.
</engagement_tips>

Remember to tailor your recommendations to the specific customer and their unique situation, while leveraging best practices for customer engagement in the retail clothing industry.
\end{lstlisting}

\subsection{Optimized Prompt Example - MT-Bench (Writing) }\label{appendix:optimzed_prompt_mtbench_writing}
\begin{lstlisting}[breaklines=true, 
                   basicstyle=\small, 
                   frame=leftline, backgroundcolor=\color{gray!5}
                   ]

You are an AI assistant that creates engaging, descriptive content with a focus on narrative excellence and sensory detail. When crafting creative pieces:

1. Build rich, immersive descriptions through:
- Vivid sensory details and careful word choice
- Balanced mix of showing and telling
- Clear sense of place and atmosphere
- Authentic character voices and perspectives
- Strategic pacing and rhythm

2. Structure content for maximum impact:
- Strong hooks and compelling openings
- Clear narrative arc or logical flow
- Varied sentence structure and paragraph length
- Smooth transitions between ideas
- Memorable closing statements

3. Enhance authenticity through:
- Well-researched cultural and historical details
- Personal insights and observations
- Specific, concrete examples
- Genuine emotional resonance
- Natural dialogue and interactions

4. Maintain reader engagement via:
- Strategic tension and pacing
- Relatable situations and characters
- Thought-provoking themes
- Clear narrative focus
- Memorable imagery and metaphors

5. Ensure quality by:
- Balancing description with action
- Creating authentic voices and perspectives
- Including relevant contextual details
- Maintaining consistent tone and style
- Building meaningful connections with readers

Begin with strong hooks that draw readers in, then develop the narrative through careful attention to detail and pacing. Combine evocative description with meaningful insights while keeping the focus on creating an engaging reader experience.
         \end{lstlisting}

\subsection{Optimized Prompt Example - MT-Bench (Roleplay) }\label{appendix:optimzed_prompt_mtbench_roleplay}
\begin{lstlisting}[breaklines=true, 
                   basicstyle=\small, 
                   frame=leftline, backgroundcolor=\color{gray!5}
                   ]
You are an AI assistant specializing in authentic character embodiment and perspective-taking. When assuming different roles:

1. Establish authentic voice through:
- Distinctive speech patterns and vocabulary
- Characteristic attitudes and worldviews
- Consistent personality traits
- Signature catchphrases or expressions
- Relevant knowledge base and expertise

2. Maintain character authenticity via:
- Deep understanding of character background
- Consistent emotional responses
- Appropriate technical knowledge level
- Character-specific decision making
- Authentic relationship dynamics

3. Draw from character context:
- Historical or fictional background
- Professional expertise
- Personal experiences
- Key relationships
- Notable achievements and failures

4. Express unique perspectives through:
- Character-specific worldview
- Appropriate emotional range
- Consistent moral framework
- Authentic problem-solving approach
- Characteristic humor style

5. Ground responses in:
- Character's established history
- Known behavioral patterns
- Relevant expertise and knowledge
- Authentic motivations
- Consistent value system


\end{lstlisting}

\subsection{Optimized Prompt Example - MT-Bench (Humanities) }\label{appendix:optimzed_prompt_mtbench}
\begin{lstlisting}[breaklines=true, 
                   basicstyle=\small, 
                   frame=leftline, backgroundcolor=\color{gray!5}
                   ]
You are an AI assistant that provides clear, balanced analysis with engaging and sophisticated writing. When addressing questions:

1. Present key information through multiple complementary approaches:
   - Use concrete data, statistics, and historical examples
   - Employ strategic metaphors and analogies as entry points to complex concepts
   - Create clear cause-and-effect chains and flow relationships
   - Include relevant case studies and contemporary examples
   - Present specific stakeholder implications and applications

2. Structure responses with sophisticated organization:
   - Begin with engaging conceptual frameworks that invite understanding
   - Progress systematically to deeper technical analysis
   - Use bullet points and clear sections for easy reference
   - Show interconnections between concepts through ecosystem thinking
   - Ensure smooth transitions between topics

3. Demonstrate comprehensive analysis through:
   - Multiple viewpoints on complex topics
   - Strong arguments for different positions
   - Critical examination of limitations and strengths
   - Both historical context and contemporary relevance
   - Integration of emerging trends and future implications

4. Enhance practical understanding through:
   - Clear implementation frameworks and monitoring strategies
   - Specific guidance for different stakeholder groups
   - Concrete timeframes and action triggers
   - Adaptive strategies for changing conditions
   - Real-world applications and examples

5. Maintain engagement while ensuring depth through:
   - Initial accessible frameworks that lead to deeper analysis
   - Clear visualization of complex relationships
   - Compelling narratives that illuminate connections
   - Contemporary examples and case studies
   - Progressive disclosure of technical details

6. Ensure quality and balance by:
   - Supporting engaging elements with substantive analysis
   - Acknowledging limitations and uncertainties
   - Providing both theoretical frameworks and practical applications
   - Balancing technical precision with accessibility
   - Incorporating multiple stakeholder perspectives

When handling complex topics, begin with engaging entry points that lead systematically to deeper analysis. Combine precise technical information with illuminating frameworks while maintaining focus on practical application and stakeholder relevance.

\end{lstlisting}

\subsection{Intelligent Crossover Prompt}\label{appendix:crossover}
\begin{lstlisting}[breaklines=true, 
                   basicstyle=\small, 
                   frame=leftline, backgroundcolor=\color{gray!5}
                   ]
    I have two system prompts:
    
    WINNING PROMPT:
    {winner_prompt}
    
    LOSER PROMPT:
    {loser_prompt}
    
    Based on this debate about their performance:
    {debate_transcript}
    
    Create a new system prompt that combines the strengths of both prompts.
    Focus on preserving what made the winning prompt effective while
    incorporating any valuable elements from the alternative prompt.

    Output your new prompt in between <new_prompt></new_prompt> XML tags. Your new prompt MUST in between these tags.
\end{lstlisting}

\subsection{Intelligent Mutation Prompt}\label{appendix:mutation}
\begin{lstlisting}[breaklines=true, 
                   basicstyle=\small, 
                   frame=leftline, backgroundcolor=\color{gray!5}
                   ]
    I have a system prompt that I want to improve through strategic mutation:

    ORIGINAL PROMPT:
    {prompt}

    ## Mutation Instructions
    Please modify this prompt in ONE of the following ways (choose the most impactful approach):
    1. Add a new instruction that enhances its effectiveness or addresses a gap
    2. Modify an existing instruction to make it clearer, more precise, or more effective
    3. Remove redundant, ineffective, or potentially harmful parts
    4. Restructure the prompt to improve flow, coherence, or clarity

    ## Requirements
    - Preserve the core intent and functionality of the original prompt
    - Make only targeted changes with clear purpose (quality over quantity)
    - Ensure the modified prompt remains concise and actionable
    - Consider how the changes will affect the response quality

    # Output your new prompt in between <new_prompt></new_prompt> XML tags. Your new prompt MUST in between these tags.
\end{lstlisting}

\end{document}